\documentclass[11pt,a4paper]{article}
\usepackage[hyperref]{emnlp-ijcnlp-2019}
\usepackage{times}
\usepackage{latexsym}

\usepackage{graphicx,graphics,wrapfig,epstopdf,subfig}
\usepackage{algorithm,algpseudocode,adjustbox}
\usepackage{url}
\usepackage{soul}

\usepackage{tcolorbox}

\newcommand{\beq}{\begin{equation}}
\newcommand{\eeq}{\end{equation}}
\newcommand{\beqs}{\begin{eqnarray}}
\newcommand{\eeqs}{\end{eqnarray}}
\newcommand{\barr}{\begin{array}}
	\newcommand{\earr}{\end{array}}

\newcommand{\eg}{\textit{e.g.}}
\newcommand{\ie}{\textit{i.e.}}

\usepackage{amsmath,amsfonts,bm}

\usepackage{multirow}

\def\eqref#1{equation~\ref{#1}}
\def\1{\bm{1}}

\def\rvu{{\mathbf{i}}}

\def\rvp{{\mathbf{p}}}

\def\rvu{{\mathbf{u}}}

\def\rvx{{\mathbf{x}}}

\def\ervp{{\textnormal{p}}}

\def\ervu{{\textnormal{u}}}

\DeclareMathAlphabet{\mathsfit}{\encodingdefault}{\sfdefault}{m}{sl}
\SetMathAlphabet{\mathsfit}{bold}{\encodingdefault}{\sfdefault}{bx}{n}

\def\sR{{\mathbb{R}}}

\def\sX{{\mathbb{X}}}

\newcommand{\E}{\mathbb{E}}

\newcommand{\R}{\mathbb{R}}

\newcommand{\Amat}{{\bf A}}

\newcommand{\Cmat}{{\bf C}}

\newcommand{\Mmat}{{\bf M}}

\newcommand{\Qmat}{{\bf Q}}

\newcommand{\Tmat}{{\bf T}}

\newcommand{\av}{{\boldsymbol a}}

\newcommand{\hv}{{\boldsymbol h}}

\newcommand{\sv}{{\boldsymbol s}}

\newcommand{\xv}{{\boldsymbol x}}
\newcommand{\yv}{{\boldsymbol y}}

\newcommand{\thetav}{{\boldsymbol \theta}}

\newcommand{\Lcal}{\mathcal{L}}

\newcommand{\Rcal}{\mathcal{R}}

\aclfinalcopy % Uncomment this line for the final submission
\def\aclpaperid{2389} %  Enter the acl Paper ID here

\title{Improving Text Generation with Student-Forcing Optimal Transport}

  \author{Guoyin Wang$^{*1}$, Chunyuan Li$^{*2}$,Jianqiao Li\thanks{* Equal contribution} $^{3}$ , Hao Fu$^3$, Yuh-Chen Lin$^1$\\
  \textbf{ Liqun Chen$^3$, Yizhe Zhang$^2$, Chenyang Tao$^3$, Ruiyi Zhang$^3$}\\ 
  \textbf{ Wenlin Wang$^3$, Dinghan Shen$^4$, Qian Yang$^3$ and Lawrence Carin$^3$} \\
  $^1$Amazon~~~~
  $^2$Microsoft Research~~~~
  $^3$Duke University~~~~
  $^4$Microsoft Dynamics 365 AI
   \\
  \texttt{\small  guoyiwan@amazon.com, chunyl@microsoft.com,jianqiao.li@duke.edu }}

\date{}

\begin{document}
\maketitle
\begin{abstract}
%Maximum likelihood estimation (MLE) is widely used in natural language generation.
Neural language models are often trained with maximum likelihood estimation (MLE), where the next word is generated conditioned on the ground-truth word tokens. During testing, however, the model is instead conditioned on previously generated tokens, resulting in what is termed {\it exposure bias}. To reduce this gap between training and testing, we propose using optimal transport (OT) to match the sequences generated in these two modes. An extension is further proposed to improve the OT learning, based on the structural and contextual information of the text sequences. The effectiveness of the proposed method is validated on machine translation, text summarization, and text generation tasks.
\end{abstract}

\section{Introduction}
\vspace{-1mm}
% The importance of text generation
% \textcolor{red}{Natural language} 
Natural language generation is an essential component  of many NLP applications, such as machine translation \citep{bahdanau2014neural}, image captioning \citep{you2016image}, text summarization \citep{see2017get}, dialogue systems \citep{vinyals2015neural}, and machine comprehension \citep{nguyen2016ms}. Generating human-like natural language is typically cast as predicting a sequence of consecutive words in a recurrent manner. Maximum likelihood estimation (MLE) is commonly employed as the objective to train such text-generation models, maximizing the log-likelihood of producing the ground-truth tokens within a sentence or paragraph \citep{salakhutdinov2015learning}. In Recurrent Neural Network (RNN) models, this is also known as Teacher-Forcing (TF) \citep{williams1989learning}, due to the use of ground-truth tokens for next-token prediction.

% Exposure bias challenge
However, in the maximum likelihood paradigm, previous observed tokens are usually provided during training, giving rise to an issue termed \emph{exposure bias} \citep{bengio2015scheduled}: the ground-truth tokens seen by the model during training are not available at inference time. During inference, the model is required to use outputs from the last step instead of the unseen ground-truth, which is often referred to as Student-Forcing (SF). As a result, there is a discrepancy between training and inference, accumulating errors along the sequence-generation trajectory \citep{Ranzato2016SequenceLT}. 

% Prior methods on solving this problem => why not ideal?
This challenge has been addressed by incorporating model-generated text at training time. \citet{bengio2015scheduled} proposed {\it scheduled sampling} (SS), where the training samples are systematically tampered with by replacing some of the ground-truth tokens with model-predicted tokens. Other works regard text generation as a sequential decision making problem, applying reinforcement learning (RL) techniques \citep{ranzato2015sequence, bahdanau2016actor}. In particular, quantitative evaluation metrics such as BLEU and ROGUE are used as sequence-level rewards for model-generated texts. 

Despite the encouraging results reported, concerns have been raised w.r.t. the above strategies. Scheduled sampling is known to be statistically inconsistent and fails to address the fundamental issues \citep{Huszar2015HowT}. RL solutions usually suffer from slow and unstable training due to the high variance of REINFORCE-based policy gradients; consequently, specific training techniques are often needed \citep{rennie2017self, liu2017action}. Additionally, the results from RL models often do not correlate well with human evaluations, as the rewards used are typically biased towards specific aspects of a language model \citep{wang2018no}. 

% However, this strategy is shown to be inconsistent and fail to address the fundamental problems \citep{Huszar2015HowT}. Another line of works leverages desired evaluation metrics and reinforcement learning (RL) \citep{ranzato2015sequence, bahdanau2016actor}, casting text generation as a sequential decision process and obtaining a sequence-level training loss between entire generated and reference sequences. These methods usually suffer from the large variance of policy-gradient estimation and requires carefully designed baselines (\eg self-critic) to make the training more robust \citep{rennie2017self,liu2017action}. Moreover, these metrics are found to be highly biased towards certain aspects and do not correlate well with human evaluations \citep{wang2018no}. 

On the other hand, recent developments in likelihood-free modeling techniques, most prominently the {\it generative adversarial networks} (GANs), tackle exposure bias in a more principled fashion \citep{lamb2016professor, yu2017seqgan,  zhang2017adversarial, li2017adversarial}. % These methods seek to match the model to the unknown real text generator w.r.t. proper discrepancy scores such as Jensen-Shannon divergence (JSD), which, unlike MLE, explicitly depend on model-generated texts. 
%Intuitively, models from this category are trained to compose texts that are indistinguishable from the real ones, judged by a dynamically adjusted critic functions (\ie, an adversarially trained discriminator) to provide sentence-level discrimination score. 
% Adversarial-based methods are also widely explored to alleviate the discrepancy between training and inference time \citep{lamb2016professor, yu2017seqgan,  zhang2017adversarial, li2017adversarial}. 
%One example of this category is the Professor Forcing algorithm, which leverages a discriminator to encourage the dynamics of generator in the training and inference mode to be the same. 
% a discriminator is typically utilized to classify the sentences sampled from the true and fake distributions.
% \yz{ideally, a discriminator is typically utilized to provide sentence-level discrimination score to guide the free-running generation toward ground-truth target. }
GAN-based text models, however, suffer from a number of severe difficulties, including mode collapse, where generated text looks real but lacks necessary diversity \citep{zhu2018texygen, caccia2018language}. Additionally, the training of GAN-based models is often unstable, and model learning easily breaks down in the event of vanishing or exploding gradients \citep{arjovsky2017wasserstein,zhang2017adversarial}. 
% Additionally, because of the lack of direct supervision from training data, adversarial training in text also requires a well-pretrained MLE model and sophisticated parameter tuning, especially when coupled with RL methods \citep{zhang2017adversarial, chen2018adversarial}. 
% Moreover, \citet{caccia2018language} recently argues that adversarial training cannot outperform MLE training when considering both quality and diversity. 
Therefore, existing adversarial methods may not be able to match sentences generated by student-forcing with ground-truth sentences. 
 
To mitigate the challenges from the adversarial methods, we utilize a sequence-matching loss based on Optimal Transport (OT), which avoids a neural discriminator and delivers a smoother gradient for the generator. Recently, \citet{chen2019improving} leverage OT loss based on a Teacher-Forcing scheme, however, it degenerates to word-level matching, making it difficult to capture temporal semantic information.
In this work, in order to enable sequence-level matching, we instead propose an OT-based sequence-level training scheme to directly optimize the discrepancy loss between the ground-truth and free-running text samples. Further, we introduced various OT cost functions for loss calculation. The significance of this work is two-fold: firstly, this approach alleviates the exposure bias, boosting model performance at the inference stage by using a sequence-level objective between free-running output and reference. Secondly, with the use of OT, this approach provides a direct objective that is easy and robust to optimize, without biasing towards a specific, manually-defined metric.

% \st{Compared with the RL and adversarial approaches above, our model directly improve the model performance at inference stage by directly semantically guiding the free-running output and our model is simpler and robust, since neither reinforce gradient nor min-max game is involved. We also extend the previous OT approach to a sequence-level matching and extend from embedding match or seq2seq to general text generation tasks.}

% \st{uses soft-argmax to get the word embeddings of MLE outputs and utilize this embeddings to align word embedding space between encoder and decoder. Then this method degenerates to word embedding alignment in between two sequences instead of a real sequence level matching. As this method degenerates to word embedding alignment,It fails to explicitly capture the temporal semantic information either.  Moreover, this method does not explicitly break the discrepancy between Teacher-Forcing and Free-Running outputs. }
 
%  \yz{I would move this part to the related work:
%  \st{Direct} Sequence matching has also been widely explored in a various machine learning tasks \citep{guu2018generating, su2015heteroscedastic, sakoe1990dynamic, graves2014towards}. Optimal Transport (OT) is first introduced to NLP to find an optimal matching of similar words between two documents\citep{kusner2015word}.
%  }

%A discriminator is typically utilized to encourage the matching of generated and ground-truth sentences. One challenge with this strategy is that the discriminator may not provide useful signal for the generator, 

% proposed method => experiments
Our work provides the following contributions: \emph{\romannumeral1}) We introduce a novel method for text generation called Student-Forcing OT (SFOT), leveraging OT loss to improve long-term sequence sampling. \emph{\romannumeral2}) A new context-preserving OT approach is proposed to effectively match a text sequence with order information. \emph{\romannumeral3}) We examine the necessity of integrating OT with Student-Forcing via Imitation Learning. \emph{\romannumeral4}) The proposed models are robust demonstrated by extensive empirical evaluations on Neural Machine Translation (NMT), Text Summarization, and Neural Text Generation (NLG).

\section{Student Forcing Optimal Transport}
%
% Similar as Professor Forcing \citep{lamb2016professor}, the basic idea of our proposed approach is 

To reduce exposure bias, the output sequences of the generator in the teacher-forcing (training) and student-forcing (inference) stages should be indistinguishable. Therefore, we propose to use OT loss to measure sequence matching distance between the two stages, in conjunction with the maximum likelihood estimate.

\subsection{Maximum Likelihood Estimate}
% \yz{Maybe change the section name to something like: "Student-forcing regularization" or "Beyond teaching-forcing MLE training?"}
%
We denote the training dataset as $N$ sequence pairs $\mathcal{D} = \{\xv_n, \yv_n\}_{n=1}^N$, with output sequence $\xv = [x_1, \cdots, x_{T}]$ and input sequence $\yv=  [y_1, \cdots, y_{T'}]$. 
Depending on the specific task, $\yv$ may have different definitions.
For {\em seq2seq} models like neural machine translation, $\yv$ represents the source sequence, conditioned on which the target sequence $\xv$ is generated. For language modeling tasks, $\yv$ is empty, and $\xv$ becomes the unconditionally generated sequence. 

To generate a text sequence, neural language models~\cite{mikolov2010recurrent} generate every token $x_t$ conditioned on the previous tokens in an auto-regressive manner:
\begin{align}
    \label{eq:seq2seq}
    \log p_{\thetav}(\xv|\yv) = \sum_{t=1}^{T}\log p_{\thetav}(x_t|x_{<t}, \yv)
\end{align}
where $\thetav$ are model parameters, and $x_{<t}$ indicates all tokens before $t$. 
Learning $\thetav$ is often performed with maximum likelihood estimation (MLE):
\begin{align}
    \Lcal_{\text{MLE}} = \E_{(\xv, \yv)\sim \mathcal{D}}[\log p_{\thetav}(\xv|\yv)]
\end{align}
To facilitate MLE training, the {\it teacher-forcing} scheme is considered. The word tokens $x_{<t}$ from the ground-truth sequence are fed into~(\ref{eq:seq2seq}) to generate the next token:
\begin{align}
    \label{eq:teacher_forcing}
    \Tilde{\xv} = [\Tilde{x}_1, \cdots, \Tilde{x}_T] ~~\text{with} ~~\Tilde{x}_t \sim  p_{\thetav}(\Tilde{x}_t|x_{<t}, \yv)
\end{align}

Learned neural language models are often evaluated using the {\it student-forcing} scheme. The previously generated word tokens of the model are conditioned to generate the next token: 
\begin{align}
    \label{eq:student_forcing}
    \hat{\xv} = [\hat{x}_1, \cdots, \hat{x}_T] ~~\text{with} ~~\hat{x}_t \sim  p_{\thetav}(\hat{x}_t|\hat{x}_{<t}, \yv)
\end{align}
%
% \yz{the below part seems to deviate the discussion -- maybe move to other part or remove?}
The difference between (\ref{eq:teacher_forcing}) and (\ref{eq:student_forcing}) reveals a gap between training and evaluation in the MLE method. To reduce the gap, a natural idea is to define a tractable function to measure the discrepancy between ground-truth and SF-generated text sequences, to regularize TF-based MLE learning. 

% In this paper, we propose to evaluate we can push sequence $\hat{\yv}$ to be closer to $\yv$.

% In more details, one can parameterize the probability of decoding each word $\yv_j$ as:
% \begin{align}
%     p(\yv_j | \yv_{<j}, \sv_j) = \text{softmax}(\Vmat \hv_j))
% \end{align}

% where $\Vmat$ is the transformation matrix that outputs a vocabulary-sized vector, usually referred as decoding embedding matrix. $\hv_j$ is the RNN hidden unit, abstractly computed as:
% \begin{align}
%     \hv_j = f(\hv_{j-1}, \sv_j)
% \end{align}
% where $f$ represents a RNN unit (\eg LSTM unit cell).

\subsection{Student Forcing Optimal Transport}
\vspace{-2mm}
\begin{figure}[t!]%\vspace{-25pt}
	\vspace{-0mm}\centering
	\begin{tabular}{c}
		\hspace{-2mm}
		\includegraphics[height=2.5cm]{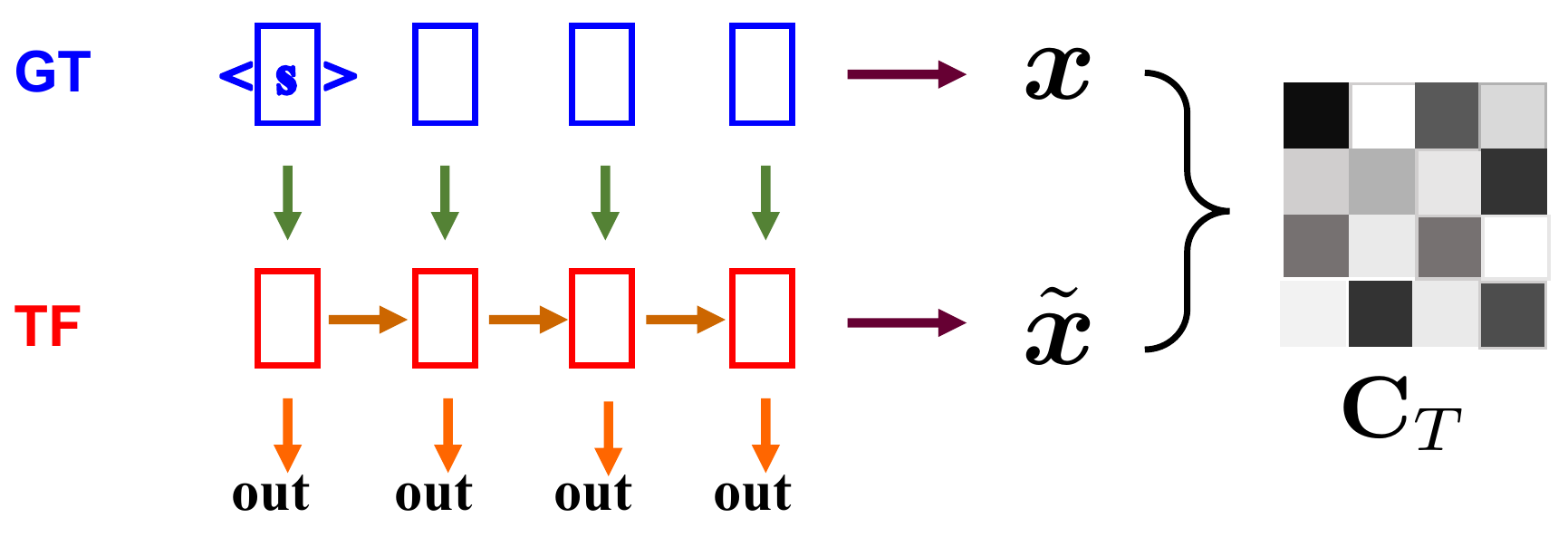} \\
		\hspace{-2mm}
		(a) Teacher-Forcing OT \vspace{2mm} \\
		% \vspace{2mm}
		\hspace{-2mm}
		\includegraphics[height=2.5cm]{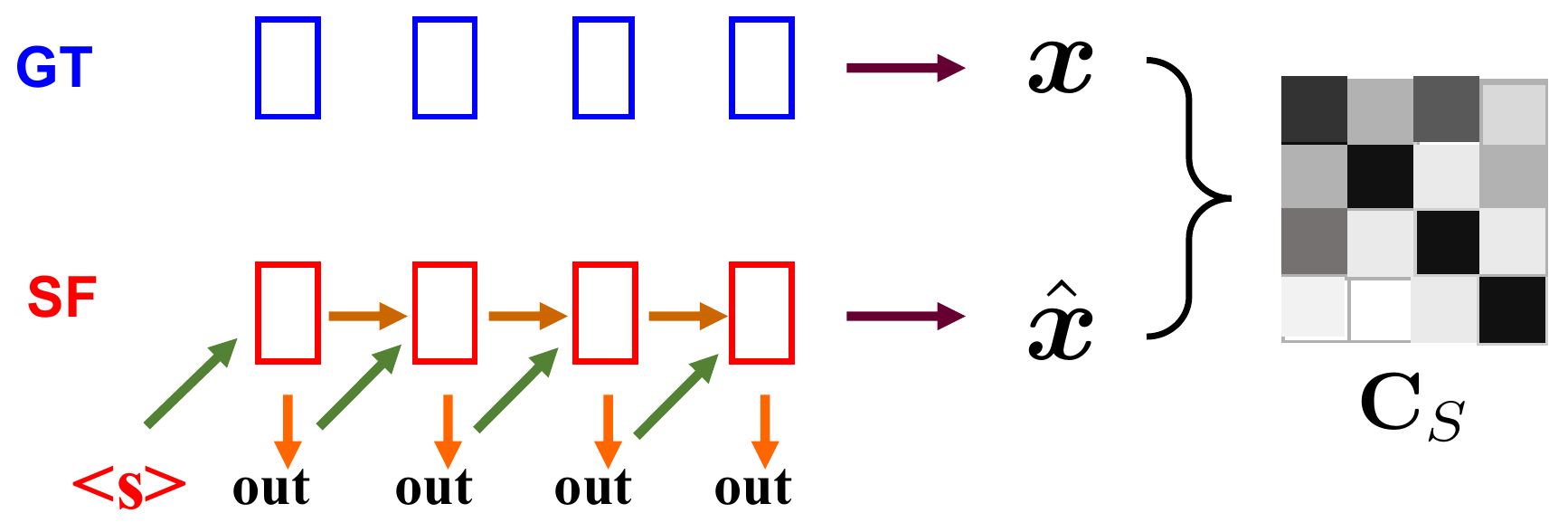} \\
		\hspace{-4mm}
		(b) Student-Forcing OT\hspace{-0mm} 
	\end{tabular}
	\vspace{-2mm}
    \caption{Comparison of (a) teacher-forcing (TF) and (b) student-forcing (SF), where $ \left \langle \text{s} \right \rangle $ is the start token.  In (a), the ground-truth (GT) sequence $\xv$ is compared with TF-generated sequence $\Tilde{\xv}$ to produce the cost matrix $\Cmat_T$; In (b), the GT sequence $\xv$ is compared with SF-generated sequence $\hat{\xv}$ to  produce the cost matrix $\Cmat_S$.}
    \label{fig:otF-framwork}
	\vspace{-4mm}
\end{figure}

% \yz{in the figure, change "Free" to "Student"? also, why the matrix in (b) looks sparse? are we claiming any sparsity for student-forcing? For the diagonalized attention matrix, we may need to explicitly elaborate this is an effect of order-preserving regularization }

We propose to use optimal transport (OT) to measure the discrepancy between the student-forcing generated sequence $\hat{\xv}$ and the ground-truth sequence $\xv$. Assuming there is an oracle/target distribution $\mu(\xv)$ to generate $\xv$, our goal is to learn $\thetav$ such that the model distribution $p_{\thetav} (\hat{\xv})$ matches $\mu(\xv)$. Formally, OT provides a distance metric between the two probability measures $\mu$ and $p$ on a domain $\mathbb{X}$ (the sequence of word tokens):
\begin{align}
\Rcal_{\mbox{ot}}(\xv, \hat{\xv}) = \inf_{\gamma \in \Pi (\mu, p)} 
\E_{(\xv, \hat{\xv} ) \sim \gamma} [c(\xv, \hat{\xv})], 
\end{align}
where $\Pi (\mu, p)$ denotes the set of all joint distributions 
$\gamma(\xv, \hat{\xv})$ with marginals $\mu(\xv)$ and $p(\hat{\xv})$.
The function $c(\xv, \hat{\xv}): \mathbb{X} \times \mathbb{X} \rightarrow \R $ defines the cost of moving $\hat{\xv}$ to $\xv$. Intuitively, OT provides a method of matching the sequence $\hat{\xv}$ to $\xv$ with the minimum cost, given $\mu$, $p$ and $c(\cdot, \cdot)$.

% During the inference phase, with same parameter $\thetav$, we can generate a free-running sequence $\hat{\xv}$, but without feeding ground-truth $\xv$ at each step. Then, the hidden unit $\hv_t$ for step $t$ is computed as 
% \begin{align}
%     \hv_t = f(\hv_{t-1}, \hat{\sv}_)
% \end{align}
% where $\hat{\sv}_j$ represents the $t$th state of the model given the observed words $\hat{\yv}_{<j}$. Each word $j$ in the synthetic sequence has the probability
% \begin{align}
%     p(\hat{\xv}_t | \hat{\xv}_{<t}, \hat{\sv}_t) = \text{softmax}(\Vmat \hv_t^f))
% \end{align}
% where $\Vmat$ is the embedding matrix and $\hv^f$ is the final layer of the model. 
% And the overall conditional probability is 
% \begin{align}
%     P_\theta(\hat{\yv}_j) = p_\theta(\hat{\yv}_j | \hat{\yv}_{< j}, \xv, \hat{\sv}_j)
% \end{align}

% The "keyword matching", refereed as {\em hard-matching} 
% \paragraph{Optimal transport background}
\paragraph{OT distance on discrete domains}
% In the paper, we mainly focus on applying the optimal transport distance on discrete data, \ie, text. 
For discrete distributions $\mu, p$ on $\sX$, we have $\mu = \sum_{i=1}^T \ervu_i \delta_{\rvx_i}$ and $p = \sum_{j=1}^{T} \ervp_j \delta_{\hat{\xv}_j}$ with $\delta_{\rvx}$ the Dirac function centered on $\rvx$. 
The weight vectors $\rvu=\{\ervu_i\}_{i=1}^T \in \Delta^T$ and $\rvp=\{\ervp_i\}_{i=1}^T \in \Delta^{T}$ belong to the $T$-dimensional simplex, \ie, $\sum_{i=1}^T \ervu_i = \sum_{j=1}^{T} \ervp_j = 1$, as both $\mu$ and $p$ are probability distributions.
Under such a setting, computing the OT distance is equivalent to solving the following network-flow problem \citep{luise2018differential}: 

\begin{align}
\vspace{-2mm}
\label{eq:ot-d}
\Rcal_{\mbox{ot}}(\xv, \hat{\xv}) &= \min_{\Mmat \in \Pi(\rvu,\rvp)}\sum^T_{i=1}\sum^T_{j=1}\Mmat_{ij} \cdot c(\xv_i,\hat{\xv}_j) \nonumber\\&= \min_{\Mmat \in \Pi(\rvu,\rvp)} \,\, \langle \Mmat, \Cmat \rangle \,
\end{align}
where $\Pi(\rvu,\rvp) = \{ \Mmat \in \sR_+^{T\times T} | \Mmat\mathbf{1}_T=\rvu, \Mmat^\top\mathbf{1}_T=\rvp \} $, $\mathbf{1}_T$ denotes a $T$-dimensional all-one vector, $\Cmat$ is the cost matrix given by $\Cmat_{ij}=c(\xv_i, \hat{\xv}_j)$, and $\langle \Mmat, \Cmat \rangle=\text{Tr}(\Mmat^\top \Cmat)$ represents the Frobenius dot-product. We refer to the minimizer $\Mmat^*$ of (\ref{eq:ot-d}) as \textit{OT matching}. 

% Comparing the two objectives, one can readily recognize that soft bipartite matching represents a special constrained solution to (\ref{eq:ot-d}), where $\Tmat$ can only take values in $\Gamma = \{ \Tmat | \max_i \{ \| \Tmat \rve_i \|_{0}, \| \rve_i^T \Tmat \|_{0} \} \leq 1, \Tmat_{ij}\in \{0,1\}, \| \Tmat \|_0 = K\}$ instead of $\Pi(\rvu,\rvv)$; here $\| \cdot \|_0$ is the $L_0$ norm and $\rve_i$ is the unit vector along $i$-th axis. 

Summarizing, our student-forcing optimal transport (SFOT) objective is:
\begin{align}
\label{eq:sfot_loss}
    \Lcal_{\text{SFOT}} = \E_{(\xv, \yv)\sim \mathcal{D}} & [\log p_{\thetav}(\xv|\yv) +  \\
    & \hspace{5mm} \lambda \E_{ \hat{\xv}\sim p_{\thetav}(\hat{\xv} | \yv)} \Rcal_{\mbox{ot}} (\xv, \hat{\xv}) ] \nonumber
\end{align}
where $\lambda $ is the weighting hyper-parameter that balances the MLE and OT losses. In practice, we only take one sample in student-forcing. Note that our SFOT objective considers $\Rcal_{\mbox{ot}} (\xv, \hat{\xv})$. This is a key difference from \citet{chen2019improving}, where teacher-forcing is used in OT $\Rcal_{\mbox{ot}} (\xv, \Tilde{\xv})$. To note the difference, we refer to the method in \citet{chen2019improving} as TFOT.

The exact minimization over $\Mmat$ is generally computationally intractable \citep{arjovsky2017wasserstein, genevay2017learning}. Hence we use the recently introduced Inexact Proximal point method for Optimal Transport (IPOT) \citep{xie2018fast} algorithm to approximate $\Mmat^*$. The details of the IPOT algorithm are shown in Appendix A.1.

\subsection{Cost Functions in SFOT}
OT-matching quality largely depends on the cost function $c(\cdot, \cdot)$.  
In particular, there is flexibility in how we represent the elements to be transported, for which we outline two alternatives below.

% To link word tokens to their semantics, we may consider various features of models as word representations. For neural language models, deep neural networks such as Long Short-Term Memory (LSTM)~\cite{hochreiter1997long} and Transformers~\cite{vaswani2017attention} are typical choices. Each word token $x_t$ is then represented as a continuous vector $\hv_t^{\ell}$, where ${\ell}$ indicates the ${\ell}$-th layer. Note that $\hv^0$ is the {\it word embedding}, which only captures the information of a single word, while $\hv^l$ for $\ell \ge 1$ often capture contextualized representations of the word in the sequence.

\paragraph{Vanilla OT}
A natural choice for the cost distance is to use the word embeddings, denoted  $\{\hv^0_t\}_{t=1}^T$, as used by previous works:
\begin{align}
    \vspace{-4mm}
    c(\xv_i, \hat{\xv}_j) = 1 - \frac{{\hv_i^0}^{\top} \hv_j^0}{\| \hv_i^0 \| \|\hv_j^0\|} \, ,
    \label{eq:sot_cost}
\end{align}

% Notice that the range of cost function $c(\cdot,\cdot) \in [0,2]$.

However, a word-embedding-based cost function only captures the token-to-token similarity. On the other hand, the semantics of words can be different in different positions or contexts. This inspires the proposal of two novel cost functions to improve text sequence matching in OT.

\begin{figure}[t]
    \centering
    \includegraphics[width = 0.45\textwidth, height=4cm]{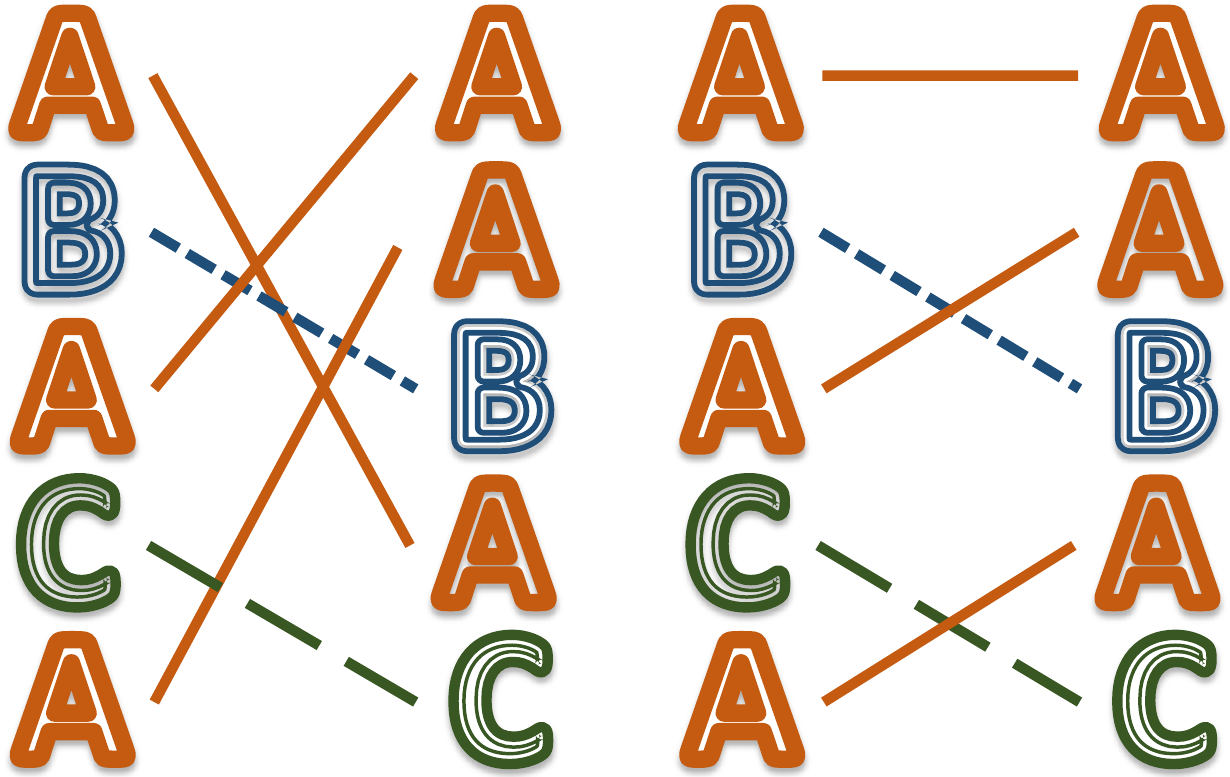}
    \caption{Illustration of the intuition of the proposed OT extensions. The goal is to match the sequence $\mathtt{ABACA}$ to the sequence $\mathtt{AABAC}$. Left: traditional OT, where only the token information is considered; thus letter $\mathtt{A}$ in any positions of the two sequences can be aligned. Right: the proposed extensions of OT, where the {\it context} and {\it ordering} information help to eliminate undesirable alignment. In terms of the context, letter $\mathtt{A}$ will be matched to those that share a similar context $\{\mathtt{A},\mathtt{B}\}$ and $\{\mathtt{A},\mathtt{C}\}$. In terms of ordering, letter $\mathtt{A}$ will be aligned to the letters with similar positions}.
    \label{fig:order-comp}
    \vspace{-6mm}
\end{figure}

\paragraph{Contextualized OT with Order-Preserving Regularizer}
Because the same word in different linguistic contexts may have different meanings, a cost function that cannot capture such variability may lead to undesirable matching results. 
While word embeddings $\{\hv^0_t\}_{t=1}^T$ may be myopic, hidden representations $\{\hv^\ell_t\}_{t=1}^T$ at higher layers ($\ell>0$) of deep language models (\eg$~$ LSTM \cite{hochreiter1997long} or the Transformer \cite{vaswani2017attention}) often capture contextualized representations of the word in the sequence.
Inspired by works on deep contextualized word representations~\citep{Peters:2018, devlin2018bert, radford2018improving}, we replace the word embeddings with $\{\hv^\ell_t\}_{t=1}^T$ to represent the meaning of words inside the sequence. Then the cost function can be defined
\vspace{-2mm}
\begin{align}
    c(\xv_i,\hat{\xv}_j) = 1 - \frac{{\hv_i^l}^{\top} \hv_j^l}{\| \hv_i^l \| \|\hv_j^l\|} %\, , \, ,
    \label{eq:cot_cost}
\end{align}
\vspace{-2mm}

Figure \ref{fig:order-comp} shows that the meaning of letter $\{\mathtt{A}\}$ can be less ambiguous when considering its local context $\{\mathtt{B,A}\}$ and $\{\mathtt{C,A}\}$. This context information further helps eliminate the undesirable matching configurations. Otherwise, letter $\mathtt{A}$ can match to $\mathtt{A}$ of any position in the other sequence. Note that contextual information may implicitly capture relatively long-term dependency information compared with vanilla OT, and it can be perceived as a ``soft" $n$-gram matching. 

We also consider an order-preserving regularizer for the contextualized OT, motivated by the fact that positional information of a token can be crucial in natural language understanding. For example, two sentences may have opposite meanings when the word order is changed: {\em ``He hated it but then started loving it''} versus {\em ``He loved it but then started hating it''.}
Hence, it is desirable to have the transportation matrix concentrate to diagonal entries, by transporting the neighboring elements in one sequence into some other neighboring elements in another sequence with nearby temporal position \citep{su2018order}. Inspired by  \citet{albregtsen2008statistical}, we penalize the contextual cost function with {\it inverse difference moment} as 
\vspace{-2mm}
\begin{align}
    c_{\text{c}}(\xv_i, \hat{\xv}_j) = c(\xv_i, \hat{\xv}_j) - \frac{\beta}{(\frac{i}{T} - \frac{j}{T})^2+1}\, 
    \label{eq:oot_cost}
\end{align}
where $\beta \ge 0$ is the weighting hyper-parameter for the order-preserving penalty. Figure \ref{fig:order-comp} shows that when only considering the token-to-token similarity, the first letter $\mathtt{A}$ from the left sequence can be moved to any $\mathtt{A}$ in the right sequence. However, when considering the ordering penalty, letter $\mathtt{A}$ is aligned to the letters with similar position. 
SFOT is summarized in Algorithm \ref{alg:FROT}.

Note that order-preserving regularization is only applied to contextualized OT, as requiring position-wise matching for the generated and target sentence may be too restrictive. Empirically, we observed adding the order-preserving regularizer to vanilla OT gives marginal improvement. However, since contextualized OT operates on a feature space, the position-wise matching is softened and can be naturally coupled with contextual cost function.

\begin{algorithm}[t]
\centering
\footnotesize
\caption{\footnotesize Student Forcing Optimal Transport}
\label{alg:FROT}
\begin{algorithmic}[1]
\State {\bfseries Input:} \footnotesize{Ground truth paired sequences $\mathcal{D}=\{\xv_n$, $\yv_n\}_{n=1}^N$ }

\State{Initialize MLE model parameters $\theta$}

\While {training}
    \State{Draw samples $\{\xv_n, \yv_n\} \sim \mathcal{D}$ }
    %and load embeddings for each token in the sample as $\{\tilde{\wv}_{n,t}\}$ and $\{\wv_{n,t}\}$}
    \State {Compute the outputs $\hv$ of the model via teacher-forcing and $\hat{\hv}$ via student-forcing}
    \State {Compute the cost matrix $\Cmat$ based on the choice of cost function in (\ref{eq:sot_cost}), (\ref{eq:oot_cost}) }
    % \If{Use contextualized embedding}
    %     \State{$\Smat_{r} = \{\tilde{\wv}_{n,t}\}$ and $\Smat_{g} = \{\hat{\hv}\}$}
    % \Elseif {}
    % \Else
    %     \State{$\Smat_{r} = \{\tilde{\wv}_{n,t}\}$ and $\Smat_{g} = \{\Vmat \text{soft-argmax} \Vmat\hat{\hv} \}$}
    % \EndIf
    % \State{Compute cost matrix $\Cmat$ as shown in \ref{eq:ot-d} or }
    \State{Compute the OT Loss $\Rcal_{\mbox{ot}}(\xv, \hat{\xv})$ defined in (\ref{eq:ot-d}) via IPOT algorithm }
    \State{Update $\theta$ by optimizing $\Lcal_{\text{SFOT}}$ defined in (\ref{eq:sfot_loss}) }
\EndWhile

\end{algorithmic}
\end{algorithm}
% \end{minipage}
% \end{wrapfigure}

\subsection{An Imitation Learning Interpretation of Student-Forcing}
The sequential text generation process can be reformulated using Markov decision processes (MDPs). Formally, an MDP $\mathcal{M} = \langle \mathcal{S}, \mathcal{A}, P_s, r \rangle$\footnote{The discount factor is set as one for simplicity} defines
a transition probability $P_s(\sv'|\sv, \av)$ from state $\sv \in \mathcal{S}$ to the next state $\sv' \in \mathcal{S}$ after the agent takes an action $\av \in \mathcal{A}$;
$r(\sv, \av)$ is an unknown reward function.
To cast text generation as an MDP, we may consider the action as the selection of the next word $x_t$ from the vocabulary, conditioned on the states of observed words $x_{<t}$, \ie, % 
% \begin{align}
  $   \sv_t = x_{<t}~~\text{and}~~\av_t = x_t $. 
% \end{align}
%
% \cl{$w_t$ is the $t$-th word token, I will define it when editing the method section}
%
At each time step $t$, the agent takes an action $\av_t$ at state $\sv_t$ according to policy $\pi(\av_t|\sv_t)$, and receives a reward $r(\sv_t, \av_t)$. 
The conjunction of the generated word and the previous methods constitute the next state $\sv_{t+1}$. 

Imitation learning seeks to learn an optimal policy from demonstrations of an expert policy $\pi_E$. In language models, the training text plays the role of expert trajectories. This objective is formally:
\begin{align} \label{eq_imitation}
\max_{r\in\mathbb{R}^{\mathcal{S}\times \mathcal{A}}} &\min_{\pi\in\Pi} ~~ -H(\pi) + \mathcal{R} (\pi_E) 
- \mathcal{R} (\pi) \\
&\text{with}~~~ 	\mathcal{R} (\pi) = \sum_{\sv, \av}\rho_{\pi}(\sv, \av)r(\sv, \av). \nonumber
\end{align}
% \vspace{-1mm}
%
where $\rho_{\pi}(\sv, \av)$ is the stationary joint distribution of $(\sv, \av)$ induced by the learning policy $\pi$; $H(\pi)$ is the entropy.  
Intuitively, the objective encourages that higher rewards are assigned to the expert policy $\pi_E$, while $\pi$ is trained to mimic $\pi_E$.

Importantly, (\ref{eq_imitation}) suggests that each individual word of sentences that induces the distribution $\rho_{\pi}$ should be fully generated by the learned language model $\pi$. 
% the sample trajectory for each occupancy measure is drawn from its corresponding policy only. 
In other words, at each time step, $\pi$ generates a word using its own previously generated words. This is exactly the student-forcing scheme we employ for the proposed SFOT algorithm.
This reveals the key difference with~\citet{chen2019improving}, where teaching-forcing is employed. Since such an agent takes actions based on the partial expert trajectories, it will induce a biased occupancy measure. This results in a sub-optimal policy for text generation, even when the imitation learning objective (\ref{eq_imitation}) is optimized.
% Hence, the performance of RL-based text generation largely depends on the design of the reward $r(\sv, \av)$.

\section{Related Work}
\vspace{-2mm}
\paragraph{Text Generation}
Natural Language Generation (NLG) is a challenging NLP task. Neural language models parameterized by autogressive architectures are widely used for NLG. To improve the global control ability of generated sentences, variational auto-encoders are considered for language generation~\cite{bowman2015generating,fu2019cyclical,fang2019implicit,li2020optimus}. Recently, GPT-2~\cite{radford2019language} and GPT-3~\cite{brown2020language} improve the generation fluency via pre-training on massive text corpus.
All of them are trained with MLE using Teacher-Forcing, which are known to suffer from {\em exposure bias} in principle\citep{bengio2015scheduled}. Several methods have been proposed to solve the problem, including~\cite{shao2018greedy,zhang2019bridging}. Adversarial training techniques were also proposed~\citep{yu2017seqgan, zhu2018texygen, che2017maximum, lin2017adversarial, guo2018long, chen2018adversarial, li2020contextualized, yang2019end, zhang2018sequence, liang2018generative}. However, adversarial-based NLG models can suffer from gradient vanishing and unstable training. Indeed, \cite{caccia2018language} argues that a temperature sweeping approach on MLE can outperform GAN-based models.
Our model further improves this work by adopting a principled sequence-matching loss via optimal transport and achieve state-of-the-art results on NLG tasks.

\vspace{-1mm}
\paragraph{Optimal Transport} 
Optimal transport is widely employed for a variety of NLP tasks, including document classification~\citep{kusner2015word}, word embedding space alignment ~\citep{alvarez2018gromov}, and generative adversarial networks~\citep{chen2018adversarial}. 
% However, different from previous works, our model takes ordering information and contextualized representations into account. The importance of word order for sequence learning is shown in Section \ref{sec:exp}.
The most related work to ours is TFOT~\citet{chen2019improving}. We discuss the difference between the proposed SFOT and TFOT as follows:

\begin{itemize}%[noitemsep,topsep=0pt]
    \item {\em Strong empirical evidence on long sentences}. While the overall performance of SFOT is superior to TFOT only by a decent margin on standard datasets, we emphasize that the main advantage of SFOT is for long sentences. As shown in the break-down analysis (Figure 4), SFOT is significantly better than TFOT when sentence length is larger than 60. This highlights the critical contribution of this paper to addressing the exposure bias problem, which longer sentences usually suffer from. 
    \vspace{-0mm}
    \item  {\em Methodology novelty}. Besides the difference in SF decoding and TF decoding in two methods, we also propose a technique on “Contextualized OT with Order-Preserving Regularizer”, which improves both SFOT and TFOT, as shown in Table 4. Note that, there is no order information used in TFOT, which degenerates it to word embedding alignment instead of sequence matching. In contrast, the proposed technique can utilize sequence level, thus improving the performance. 
    \item  {\em Theoretical difference}.
     In Section 2.3, we provide theoretical justification on why SFOT can reduce exposure bias, while TFOT still suffers from it: TFOT is based on partial expert trajectories and induces a bias occupancy measure, while our proposed method SFOT uses previous self-generated words and can obtain an optimal policy. 
    \vspace{-0mm}
\end{itemize}

\vspace{-1mm}
\paragraph{Sequence matching}
Direct sequence matching has been explored widely in various machine learning tasks. Jaccard distance has been used to retrieve prototypes for sentence generation \citep{guu2018generating}. Chernoff distance has been applied to image classification \citep{su2015heteroscedastic}. However, these distances assume each instance in the sequence is independent, ignoring temporal information. These distances consequently measure sequence alignment poorly, as they miss semantic relationships inside a sentence (\eg cause and effect). Dynamic time warping (DTW) \citep{sakoe1990dynamic} and Connectionist Temporal Classification (CTC) \citep{graves2014towards} consider temporal information, and have been adopted widely in speech recognition. However, these losses preserve strict ordered alignment, and hence cannot be directly applied to text sequences. 
%\eg inverted sentences cannot match under DTW distance.

% \begin{table}[!t] %\small
%     \centering
%     \begin{tabular}{|c|c|c|c|}
%         \hline
%         Task & Algorithm & NT2012 & NT2013 \\
%          \hline
%          \hline
%          \multirow{3}{*}{VI-EN}
%          & MLE & 21.8 & 24.5 \\
%          & TFOT  & 21.9 & 25.5 \\
%         %  VI-EN: TFOT-o & 21.9 & 25.6 \\
%          & SFOT  & \textbf{22.3} & \textbf{25.8} \\
%          \hline
%          \multirow{3}{*}{EN-VI}
%          & MLE & 23.8 & 26.1 \\
%          & TFOT & 24.5 & 26.9 \\
%         %  EN-VI: TFOT-o & 24.7 & 27.2\\
%          & SFOT & \textbf{24.9} & \textbf{27.4} \\
%          \hline
%     \end{tabular}
%     \caption{BLEU scores on VI-EN and EN-VI on pre-trained GNMT model.}
%     \label{tab:Bleu VI-EN}
% \end{table}

% \begin{table}[t!] %\small
%     \centering
%     \begin{tabular}{|c|c|c|c|}
%         \hline
        
%          Task & Algorithm &   NT2013 & NT2015 \\
%          \hline
%          \hline
%         \multirow{3}{*}{DE-EN}
%          & MLE & 29.0 & 29.9 \\
%          & TFOT  & 29.2 & 30.1 \\
%          & SFOT  & \textbf{29.3} & \textbf{30.3} \\
%          \hline
%          \multirow{3}{*}{EN-DE}
%          & MLE & 24.3 & 26.5 \\
%          & TFOT & 24.6 & 26.8 \\
%          & SFOT & \textbf{24.8} & \textbf{27.0} \\
%          \hline
%     \end{tabular}
%     \caption{BLEU scores on DE-EN and EN-DE on pre-trained GNMT model.}
%     \label{tab:Bleu DE-EN}
%     \vspace{-4mm}
% \end{table}

\vspace{-2mm}
\section{Experiments} \label{sec:exp}
\vspace{-2mm}
We perform experiments on neural machine translation (NMT), abstractive text summarization and unconditional natural language generation (NLG) tasks. Algorithms are implemented in Tensorflow and trained on an NVIDIA TITAN X GPU.

%  for small tasks and NVIDIA V100 GPU for larger tasks.

\subsection{Neural Machine Translation}
\vspace{-1mm}
Two standard datasets are tested for NMT tasks: a small-scale English-Vietnamese corpus from the IWSLT 2015 Evaluation Campaign \citep{cettolo2015iwslt} and a large-scale English-German corpus from the WMT16 Evaluate Campaign\footnote{\url{http://statmt.org/wmt16}}. Further details of the datasets and the experimental setup are shown in Appendix \ref{ap_subsec: NMT}.

\begin{table}[!t] \small
    \centering
    % \hspace{-5mm}
    \begin{tabular}{|c|c|c|c|}
        \hline
        Task & Algorithm & NT2012 & NT2013 \\
         \hline
         \hline
         \multirow{7}{*}{VI-EN}
         & MLE & 21.8 & 24.5 \\
         & SS & 21.8 & 24.6 \\
         & RAML & 22.0 & 25.0 \\
         & MIXER & 21.9 & 24.7 \\
         & SPG & 22.0 & 25.1 \\
         & TFOT  & 21.9 & 25.5 \\
        %  VI-EN: TFOT-o & 21.9 & 25.6 \\
        \cline{2-4}
         & SFOT  & \textbf{22.3} & \textbf{25.8} \\
         \hline
         \hline
         \multirow{7}{*}{EN-VI}
         & MLE & 23.8 & 26.1 \\
         & SS & 23.9 & 26.2 \\
         & RAML & 24.2 & 26.6 \\
         & MIXER & 24.0 & 26.3 \\
         & SPG & 24.3 & 26.7 \\
         & TFOT & 24.5 & 26.9 \\
        %  EN-VI: TFOT-o & 24.7 & 27.2\\
        \cline{2-4}
         & SFOT & \textbf{24.9} & \textbf{27.4} \\
         \hline
    \end{tabular}
    %  \vspace{-5pt}
    \caption{VI-EN and EN-VI translation BLEU scores.}
    %  \vspace{-5pt}
    \label{tab:Bleu VI-EN}
\end{table}
 \begin{figure}[t]
    \centering
    \hspace{-5mm}
    \includegraphics[width=.45\textwidth]{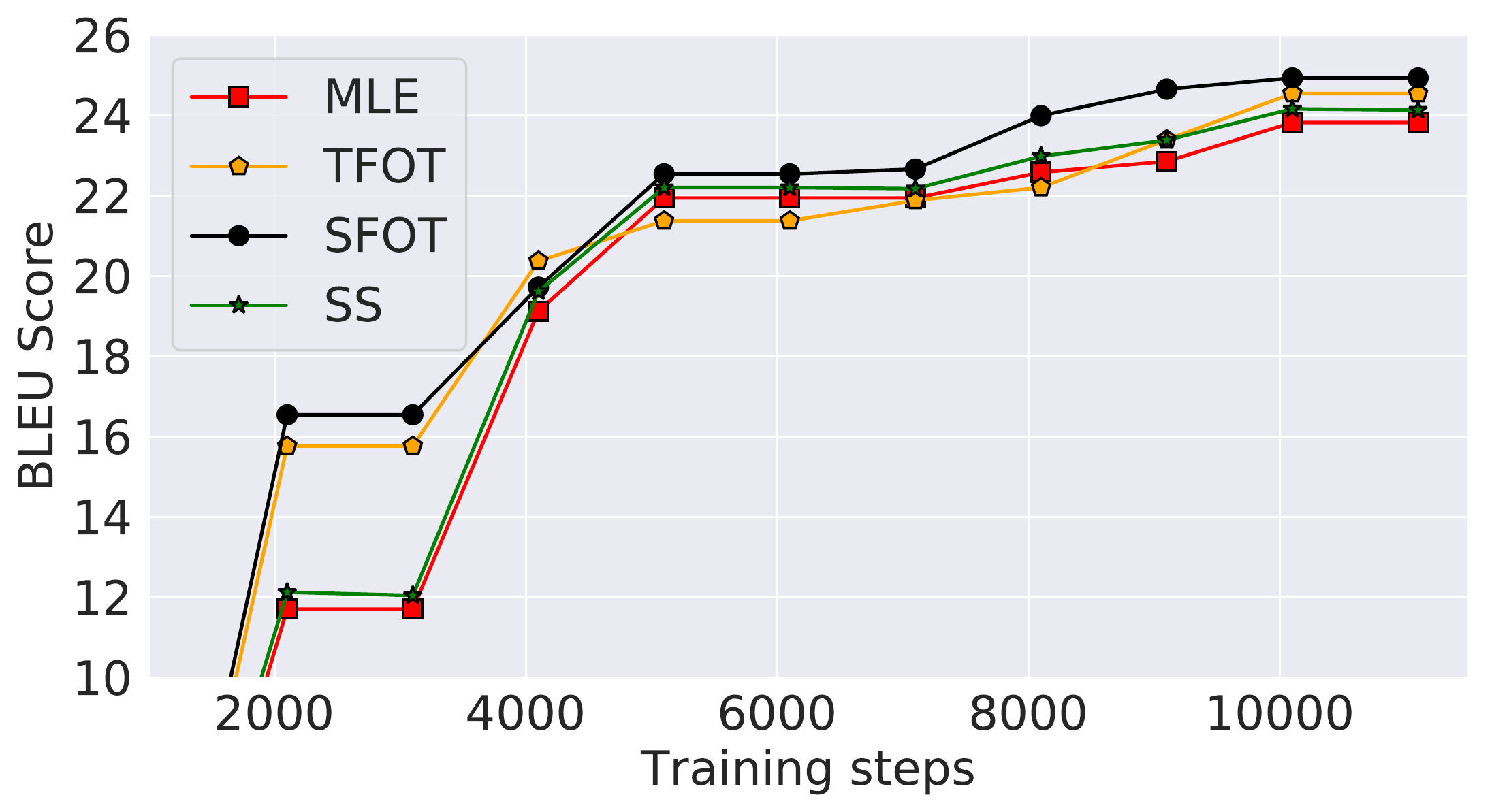}
    \caption{Convergence on NT2012 EN-VI.}
    \label{fig:convergence}
    \vspace{-4mm}
\end{figure}

 We compare SFOT with a variety of methods: MLE \citep{luong17}, Scheudle Sampling (SS), TFOT and several RL-based models, \ie~RAML \citep{norouzi2016reward}, SPG \citep{ding2017cold}, and MIXER \citep{ranzato2015sequence}. The results are summarized in Tables \ref{tab:Bleu VI-EN} and \ref{tab:Bleu DE-EN}. The proposed SFOT approach consistently improves upon MLE training and outperforms other models in all experimental setups. Besides the quantitative results, we observe that SFOT correctly maintains the information from the source side to make correct translations (Table \ref{tab:translation_example}). As can be seen from these examples, our model can better preserve information from the source and is less likely to transfer words incorrectly. Notice that most errors in the baseline models occur in the latter part of sequences, due to error accumulation from {\em exposure-bias}, which SFOT addresses by matching the free-running outputs to the ground-truth. In conjunction with the quantitative results presented above, these qualitative observations confirm that our model can generate more reliable translation for long sentences and address the {\em exposure-bias} problem.

 \begin{table}[!t] \small
    \centering
    % \hspace{-5mm}
    \begin{tabular}{|c|c|c|c|}
        \hline
        
         Task & Algorithm &   NT2013 & NT2015 \\
         \hline
         \hline
        \multirow{7}{*}{DE-EN}
         & MLE & 29.0 & 29.9 \\
         & SS & 29.0 & 29.9 \\
         & RAML & 29.1 & 30.1 \\
         & MIXER & 29.0 & 30.0 \\
         & SPG & 29.1 & 30.1 \\
         & TFOT  & 29.2 & 30.1 \\
         \cline{2-4}
         & SFOT  & \textbf{29.3} & \textbf{30.3} \\
         \hline
         \hline
         \multirow{7}{*}{EN-DE}
         & MLE & 24.3 & 26.5 \\
         & SS & 24.3 & 26.5 \\
         & RAML & 24.5 & 26.7 \\
         & MIXER & 24.4 & 26.6 \\
         & SPG & 24.5 & 26.7 \\
         & TFOT & 24.6 & 26.8 \\
         \cline{2-4}
         & SFOT & \textbf{24.8} & \textbf{27.0} \\
         \hline
    \end{tabular}
    \caption{DE-EN and EN-DE translation BLEU scores.}
    \label{tab:Bleu DE-EN}
    \vspace{-4mm}
\end{table}
 
\begin{table*} [!ht] \scriptsize 
	\centering
	% \vskip 0.0in
	% \def\arraystretch{1.0}
	% \begin{scriptsize}
	\begin{tcolorbox} 
	\hspace{-5mm}
	\begin{tabular}{c l}
% 		\toprule[1.2pt]
        {\bf Reference}: & India’s new prime minister, Narendra Modi, is meeting his Japanese counterpart, Shinzo Abe, in Tokyo to discuss economic and 
        security \\ & ties, on his first major foreign visit \textcolor{blue}{since} \textcolor{red}{winning} \textcolor{blue}{May’s election}. 
         \vspace{1mm} \\ 
        \hline  \vspace{0mm}  \\
        {\bf MLE}: & India ‘ s new prime minister , Narendra Modi , meets his Japanese counterpart , Shinzo Abe , in Tokyo , during his first major 
        foreign visit \\ & \textcolor{blue}{in May} to discuss economic and security relations . \\
        {\bf TFOT}: & India ‘ s new prime minister Narendra Modi meets his Japanese counterpart, Shinzo Abe, in Tokyo  at his first major foreign visit
        \textcolor{blue}{since his} \\ & \textcolor{blue}{election in May} in order to discuss economic and security relations . \\
        {\bf SFOT}: & India ’ s new prime minister , Narendra Modi , is meeting his Japanese counterpart Shinzo Abe in Tokyo in his first major foreign visit
         \textcolor{blue}{since} \\ & \textcolor{blue}{his election} \textcolor{red}{victory} \textcolor{blue}{in May} to discuss economic and security relations.
      \vspace{-1mm}
      \end{tabular} 
      \vspace{-1mm}
      	\end{tcolorbox}
      \vspace{-1mm}
       \begin{tcolorbox} 
      \hspace{-5mm}
    \begin{tabular}{c l}    
        {\bf Reference}: & Chinese leaders presented the Sunday ruling as a democratic breakthrough because it gives Hong Kongers a direct vote, but the 
        decision \\ & also  makes clear that Chinese leaders would retain a firm hold on the process through a nominating committee tightly 
        controlled by \textcolor{blue}{Beijing}. \vspace{1mm} \\ 
        \hline  \vspace{0mm}  \\
        {\bf MLE}: & The Chinese leadership presented the decision of Sunday as a democratic breakthrough , because it gives Hong Kong citizens a 
        direct  right \\ & to vote , but the decision also makes it clear that the Chinese leadership maintains the \textbf{expiration of} a nomination 
        committee closely \\ & controlled by \textcolor{blue}{Beijing} .\\
        
        {\bf TFOT}: & The Chinese leadership presented Sunday ' s decision as a democratic breakthrough because it gives the citizens of Hong Kong 
        a direct \\ & right to vote , but the decision also makes it clear that the Chinese leadership keeps the process firmly in the hands of a 
         \textcolor{blue}{government}\\ & -controlled  Nomination Committee. \\
        {\bf SFOT}: & The Chinese leadership presented the decision on Sunday as a democratic breakthrough , because Hong Kong citizens have a 
        direct electoral \\ & right , but the decision also makes it clear that the Chinese leadership remains firmly in hand with a nominating 
        committee controlled  by \\ & \textcolor{blue}{Beijing}.
% 		\bottomrule[1.2pt]
    \vspace{-1mm}
	\end{tabular}
	\vspace{-1mm}

	\end{tcolorbox}
	\vspace{-0mm}
	\caption{Comparison of German-to-Enlish translation examples. For each example, we show the human translation (reference) and the translation from MLE, TFOT, and SFOT. We highlight the key phrase differences between reference and translation outputs in blue and red, and annotate translation errors in bold.  In the first example, SFOT correctly maintains all the information in {\em ``since winning in May election"} by translating to {\em ``since his election victory in May"}, whereras MLE only generates {\em ``in May"} and TFOT also misses {\em ``winning"} in the reference. In the second example, SFOT successfully keeps the information {\em ``Beijing"}, whereas MLE generates wrong words {\em ``expiration of"} and TFOT changes {\em ``Beijing"} to {\em ``government"}.}
	\label{tab:translation_example}
	% \end{scriptsize}
	\vspace{-0mm}
\end{table*}

\begin{figure}[t]
    \centering
    \includegraphics[width=.45\textwidth]{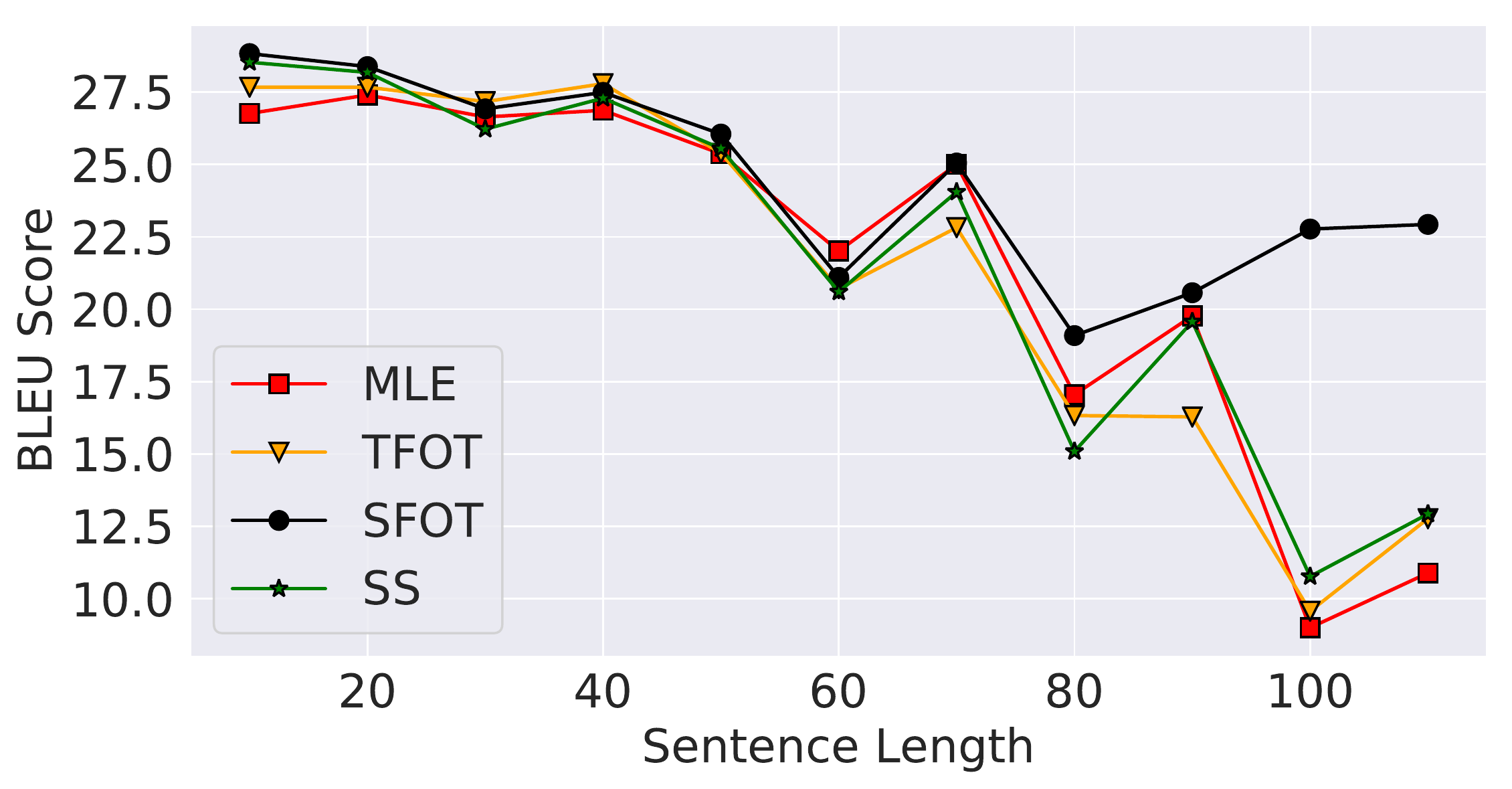}
    \caption{Translation quality as sentences become longer on NT2013 EN-VI.}
    \label{fig:bleu_seqlen}
    \vspace{-0mm}
\end{figure}

% \vspace{-2mm}
% \paragraph{Analysis} 

We further investigate the performance on the English-Vietnamese dataset. We first check the convergence of the BLEU score on the validation set, shown in Figure \ref{fig:convergence}. SS converges similarly to the MLE model and only shows marginal improvement. The models with OT converge much faster in the early stages and finish with a higher final performance. However, TFOT has an unstable convergence trajectory, as it can degrade MLE training performance. SFOT by contrast is consistently better than MLE training, achieving the best BLEU score among the three models. 

To further demonstrate SFOT's ability to address {\em exposure bias}, we follow \citet{bahdanau2014neural} and group sentences of similar lengths together, computing a BLEU score per group on the test set. As errors accumulate in generation, longer sentences suffer more from {\em exposure bias} and have lower quality. Figure \ref{fig:bleu_seqlen} shows that our model is more effective in handling long sentences: compared with other methods, SFOT is more robust for longer sentence lengths, indicating that matching the generated and ground-truth sentence in the sequence level may alleviate {\em exposure bias} for long-text generation. 

Additionally, we compare different OT cost function variants in Table \ref{tab:Bleu VI-EN-e2e}. We denote contextualized OT with order-preserving regularizer as -c. Contextualized OT improves translation quality when used with TFOT and SFOT. The improvements on both models indicate that contextualized representations with order preservation is capable of capturing more sentence semantic information.

% To better observe the change in training, we plot the training curve as shown in Figure \ref{fig:convergence} for all the considered methods. As FROT-oh and FROT-o are pretty close, we only show the FROT-o in the plot. It is obviously that OT-based method can greatly speed up the training procedure in the early stage of training. Another finding is that, FROT and FROT-o are consistently better than the MLE training, while TFOT can reduce the translation quality during the latter training stage. 

\begin{table}[!t] \small
    \centering
    \begin{tabular}{|c|c|c|c|}
        \hline
         Task & Algorithm & NT2012 & NT2013 \\
         \hline \hline
         \multirow{5}{*}{VI-EN}
         & MLE & 21.8 & 24.5 \\
         & TFOT  & 22.2 & 25.1 \\
        %  & TFOT-o & 22.0 & 24.7\\
         & TFOT-c & 22.2  & 25.4\\
         & SFOT & 22.2 & 25.6 \\
        %  & SFOT-o  & \textbf{22.3} & 25.1 \\
         & SFOT-c & \textbf{22.3} & \textbf{25.8} \\
         \hline
         \multirow{5}{*}{EN-VI}
         & MLE & 23.8 & 26.1 \\
         & TFOT & 24.5 & 26.9 \\
        %  & TFOT-o & 24.3 & 26.6 \\
         & TFOT-c & 24.8 & 27.2 \\
         & SFOT & 24.8 & \textbf{27.4} \\
        %  & SFOT-o & \textbf{24.4} & \textbf{26.8} \\
         & SFOT-c & \textbf{24.9} & \textbf{27.4} \\
         \hline
    \end{tabular}
    \caption{BLEU scores for VI-EN and EN-VI ablation study.}
    \label{tab:Bleu VI-EN-e2e}
    \vspace{-0mm}
\end{table}

% MARK
% In Table \ref{tab:translation_example}, we show sample translations from German to English for qualitative assessment. The main differences between references and the translation outputs are highlighted in color. We observe that SFOT model correctly maintains the information from the source side to make correct translation. In the first example, SFOT correctly maintains all the information in {\em "since winning in May election"} by translating to {\em "since his election victory in May"}, whereras GNMT only generate {\em "in May"} and TFOT also misses the {\em "winning"} in the reference. In the second example, SFOT successfully keeps the information {\em "Beijing"}, whereas GNMT generates wrong words {\em "expiration of"} and TFOT changes {\em "Beijing"} to {\em "government"}. These examples show that our model are more precise to keep the information from the source, and less likely to transfer words to wrong words or weakly related words. In conjunction with the quantitative results presented above, these qualitative observations confirm that our model can generate more reliable translation for the long sentences, and address the {\em exposure-bias} problem. 

\subsection{Abstractive Text Summarization}
\vspace{-1mm}
We use a widely considered English Gigawords corpus \citep{graff2003english} for the text-summarization task. Similar to NMT experiments, we use MLE as our baseline model and further compare SFOT with SS, TFOT, and several RL-based methods, \ie~RAML, SPG and MIXER. We evaluate the model performance using ROUGE (including -1, -2, -L) score \citep{lin2004rouge}, the most popular metric for summarization. Details of the datsetsets and the experimental setup are shown in Appendix \ref{ap_subsec:TS}.

% \paragraph{Dataset} We use a widely accepted English Gigawords corpus \citep{graff2003english} for the text summarization task. We follow the pre-process in \citep{rush2015neural}. The dataset is sampled and splitted into train/dev/test set with size 200K/8K/2K. 

% \paragraph{Setup} Similar to NMT experiments, we use MLE as our baseline model and further compare SFOT with SS, TFOT and several RL-based methods, \ie RAML, SPG and MIXER. We evaluate the model performance using most ROUGE (including -1, -2, -L) score \citep{lin2004rouge}, which is the most popular metric for summarization. \gw{further details are needed here}

% Table \ref{tab:giga} shows the summarization results. SFOT outperforms all other methods by a reasonable margin, which is consistent with our NMT results. The improvement in ROUGE score indicates the contextualized matching is beneficial to generation quality. SFOT outperforming the RL-based models indicates that OT matching performs better than direct word/phrase matching in RL-rewards.
Summarization results are provided in Table \ref{tab:giga}. Consistent with our NMT results, SFOT outperforms all other methods, showing that the contextualized matching is capable of capturing semantic information essential for high-quality generation. Moreover, the superiority of SFOT over RL-based models demonstrates that the OT (sentence) matching is more robust than word/phrase matching in RL rewards.
\begin{table}[t!]
    \centering
    \small
    \begin{tabular}{|c| c| c| c|} %{L{5.5cm} l}
    \hline
    
    { Method} & { ROUGE-1} & { ROUGE-2} & { ROUGE-L}\\ 
    \hline
    \hline
    MLE & $36.1 $ & $16.4 $ & $32.3 $ \\
    SS & $36.6 $ & $16.8 $ & $32.7 $ \\
    RAML & $36.3 $ & $16.7 $ & $32.5 $ \\
    SPG & $36.5 $ & $16.8 $ & $32.8 $ \\
    MIXER & $36.3 $ & $16.6 $ & $32.6 $ \\

    % ERPO\cite{tan2018connecting} & $36.72 \pm 0.29$ & $16.99 \pm 0.17$ & $32.95 \pm 0.33$ \\
    TFOT & $36.8 $ & $17.2 $ & $33.5 $ \\
    \hline
    SFOT & \bm{$37.0 $} & \bm{$17.5 $} & \bm{$33.9$} \\
    \hline
    \end{tabular}
    \vspace{-5pt}
    \caption{Results of text summarization on English Gigawords dataset.}
    \label{tab:giga}
    \vspace{-0pt}
    \vspace{-0mm}
\end{table}

\subsection{Neural Language Generation}
\vspace{-2mm}
% \paragraph{Dataset} 
Following recent unconditional long text generation work~\citep{caccia2018language}, we perform experiments on EMNLP2017 WMT News dataset\footnote{\url{http://www.statmt.org/wmt17/}}. %200K sentences are collected as the training set and 10K sentences as the test set. 
All sentences are longer than 20, making the dataset appropriate for testing the {\em exposure bias} problem. % is more sever for this task.

% \paragraph{Setup} 
To evaluate the effectiveness of our model, we consider various baseline methods, including recent GAN-based text generation approaches, such as SeqGAN~\citep{ramachandran2016unsupervised}, RankGAN~\citep{lin2017adversarial}, MaliGAN~\citep{che2017maximum}, and LeakGAN~\citep{guo2018long}, as well as an MLE-trained model using temperature sweep~\citep{caccia2018language}. We apply SFOT with contextualized OT to improve the temperature-sweep MLE model. SS does not show significantly different results compared to the MLE model. 
%More details of experiment setup are shown in Appendix \ref{subsec:NLG_ex_setup}.
% In the experiment, we set OT weighting parameter $\lambda=1$ and the order-preserving penalty weighting parameter is $\beta=0.1$. Since input sequence $\yv$ is empty in a language model, to better guide the student forcing output, we adopt schedule sampling with ratio $0.3$ in our experiments. 
For the evaluation metric, we follow the current protocol for NLG evaluation~\cite{zhu2018texygen} w.r.t. both quality and diversity. Specifically, the quality of the generation is measured with BLEU score~\cite{papineni2002bleu} and the diversity is evaluated with Self-BLEU~\cite{zhu2018texygen}. Human evaluation is further considered to measure the quality of generation\footnote{We perform human evaluation using Amazon Mechanical Turk. 100 generated sentences are sampled from each model. Ten native speakers are asked to rate each sentence in the scale 1 to 5 in terms of readability and meaningfulness}. More details of experiment setup are shown in Appendix \ref{ap_subsec:NLG}.   

% We examine several recent GAN text generation models and a MLE model with temperature swapping \citep{caccia2018language} as our baseline model. We then perform SFOT on top of the MLE model by introducing a OT sequence matching between ground-truth and student forcing outputs with schedule sampling ratio $0.3$. We follow the current protocol for NLG evaluation with respect to both quality and diversity. We evaluate quality of generation with BLEU score \citep{papineni2002bleu} and evaluate diversity with Self-BLEU \citep{zhu2018texygen}.

To reasonably select the best model along the temperature sweep, we are motivated by~\cite{gu2018dialogwae} and propose the BLEU-F1 score to evaluate the trade-off between the quality and diversity simultaneously, defined as
% We use BLEU-F1 score to evaluate quality and diversity trade-off~ \citet{gu2018dialogwae}. BLEU-F1 score is defined as the geometry average of BLEU score and $1-$ Self-BLEU
\begin{align} \small
    \text{BLEU-F1} = \frac{2 \times \text{BLEU}\times (\text{1-Self-BLEU})}{\text{BLEU}+(\text{1-Self-BLEU})}\, .
\end{align}

\begin{figure}[t]
    \vspace{-0mm}
    \centering
    \includegraphics[width=.45\textwidth]{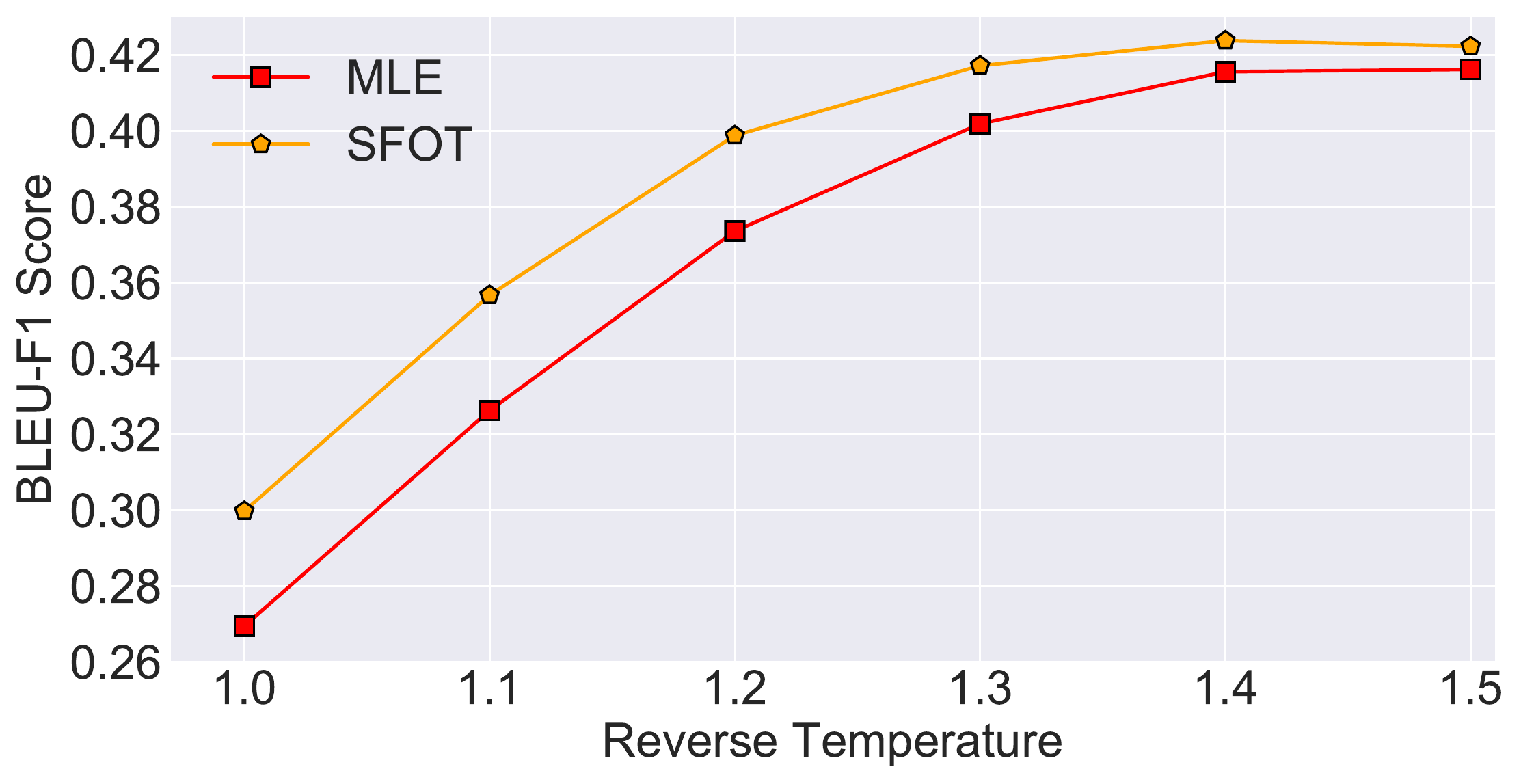}
    \caption{BLEU-5 F1 score plotted against reverse temperature $\alpha$ on EMNLP2017 News test set.}
    \label{fig:news_bleu_F1}
    \vspace{-0mm}
\end{figure}

Figure~\ref{fig:news_bleu_F1} shows the BLEU-F1 score versus reverse temperature on MLE and SFOT. We observed that the best temperature for MLE model is $1/1.5$ and for SFOT is $1/1.4$. We further conduct analysis under these temperatures. Figure~\ref{fig:news_bleu_F1} also indicates that the SFOT model consistently improves the MLE model on the BLEU-F1 score. 

\begin{figure}[t]
    \vspace{-0mm}
    \centering
    \includegraphics[width=.45\textwidth]{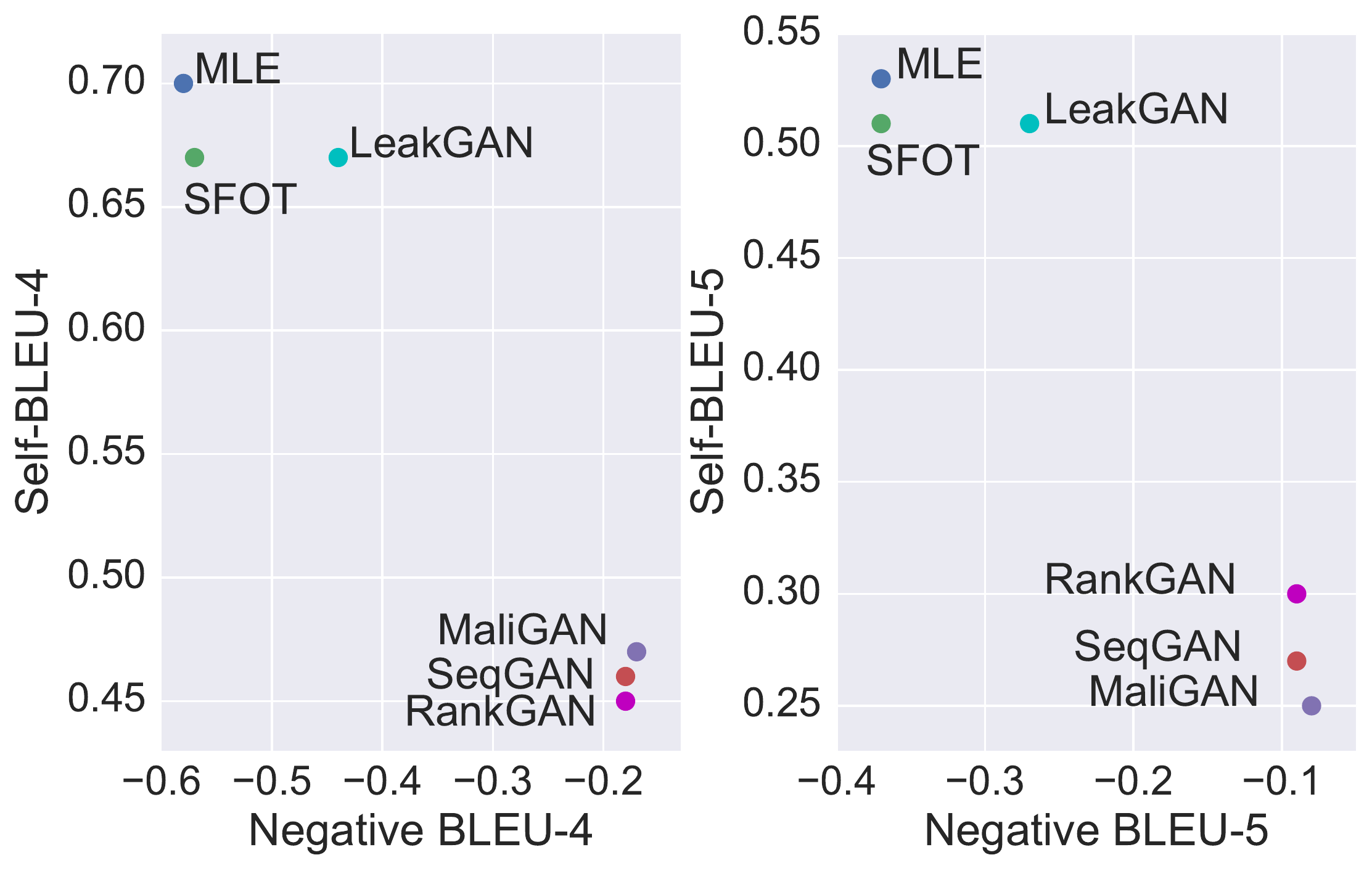}
    \caption{Negative BLEU (lower is better) plotted against Self-BLEU (lower is better) for EMNLP2017 News test set, for BLEU-4 (left) and BLEU-5 (right).}
    % \label{fig:news_bleu_F1}
    \label{fig:news_bleu}
    \vspace{-0mm}
\end{figure}

We compare SFOT with the proposed strong baselines in Figure~\ref{fig:news_bleu} and report human evaluation of generated quality in Table~\ref{tab:NLG_eval}. We observe that SFOT has the highest generation quality in human evaluation.  
%We observe that under the same quality of generation, SFOT greatly improves MLE mode in diversity. 
With better guided sequence-level semantics information, SFOT generates high-quality sentences at higher temperatures, compared with the MLE model. The MLE model decreases temperature to concentrate on generating {\em safe} words (with high probability); this avoids error accumulation by avoiding risky words. However, the model loses generation diversity when increasing temperature, as the model only focuses on {\em safe} words. SFOT obtains better quality at higher temperature, indicating SFOT can generate reasonable sequences on more risky words and hence can address {\em exposure bias} and gain more diversity in generation.
%By decreasing temperature, the model gains more diversity in generation.
We also observe that SFOT outperforms all text GANs in terms of quality-diversity trade-off and human evaluation. Under similar Self-BLEU score, SFOT significantly improves the quality of LeakGAN~\cite{guo2018long}, the best GAN by BLEU metric.

\begin{table}[t!]
    \centering
    \small
    \begin{tabular}{|c| c| c| c| c|} %{L{5.5cm} l}
    \hline
    
    { Method} & { SeqGAN} & { RankGAN} & { MaliGAN} \\ 
    \hline
    Score & $2.56 \pm 0.49$ & $ 2.89 \pm 0.56$ & $2.50 \pm 0.46$ \\
    \hline 
    \hline
    { Method}  & LeakGAN & { MLE} & { SFOT}  \\ 
    \hline
    Score  &$3.45 \pm 0.47$ & $ 3.46 \pm 0.42$ & $ 3.54 \pm 0.37$ \\
    \hline
    \end{tabular}
    % \vspace{-5pt}
    \caption{Human evaluation of NLG on EMNLP news 2017 dataset. 100 generated sentences from each model are rated 1-5, with means and standard deviations reported. Real sentences were rated $4.21 \pm 0.44$.}
    \label{tab:NLG_eval}
    \vspace{-0pt}
\end{table}
%\paragraph{Analysis} 

\section{Conclusions}
\vspace{-2mm}
We have introduced SFOT to mitigate exposure bias in text generation. 
The proposed model captures positional and contextual information of word tokens in OT matching.
Experiments on neural machine translation, text summarization, and text generation have demonstrated the effectiveness of our SFOT algorithm, yielding improved performance over strong baselines on these tasks.

% \section*{Acknowledgments}

% \newpage 

\bibliography{acl2019}
\bibliographystyle{acl_natbib}
% \clearpage
% \newpage
\appendix

\section{Appendix}
\label{sec:appendix}

\subsection{Inexact Proximal point method for Optimal Transport (IPOT) Algorithm}
\label{subsec:IPOT}

The IPOT algorithm to approximate $\Mmat^*$ is shown in Algorithm \ref{alg:ipot}
% \begin{wrapfigure}[12]{R}{0.45\textwidth}
% \begin{minipage}{0.5\textwidth}
% \vspace{-12mm}

% \begin{minipage}{1.0\textwidth}
% \begin{wrapfigure}[12]{R}{0.45\textwidth}
% \vspace{-2em}
\begin{algorithm}[H]
\footnotesize
\caption{\footnotesize IPOT algorithm}
\label{alg:ipot}
\begin{algorithmic}[1]
\State {\bfseries Input:} \footnotesize{two sequence $\xv$, $\hat{\xv}$ with length $T$, and generalized step size $1/\varepsilon$, $\sigma$ }

\State $\boldsymbol{\sigma}=\frac{1}{T^{''}}\mathbf{1_{T}}\,,$ $\Mmat^{(1)} = \mathbf{1_T} \mathbf{1_{T}}^\top\,,$

\State $\Cmat_{ij} = c(\xv_i, \hat{\xv}_j)$,  $\Amat_{ij} = {\rm e}^{-\frac{\Cmat_{ij}}{\varepsilon}}$  
% \textbf{Cost}(\Xmat, \Ymat)\,,$ $\Emat = [\frac{1}{(\frac{i}{N} - \frac{i}{M})^2+1}]_{ij}$,% $\Fmat=[l^2(i,j)]_{ij}$
% \State  $\hat{\Cmat} =  \Cmat - \lambda \Emat\,,$ $\Amat_{ij} = {\rm e}^{-\frac{\hat{\Cmat}_{ij}}{\beta}}$
\For{$t=1,2,3\ldots$}
    \State $\Qmat = \Amat \odot \Tmat^{(t)}$ \footnotesize{// $\odot$ is Hadamard product}
    \For{$k=1,\ldots K$} \footnotesize{// $K=1$ in practice}
        \State $\boldsymbol{\delta} = \frac{1}{T\Qmat{\boldsymbol{\sigma}}}$, $\boldsymbol{\sigma} = \frac{1}{T\Qmat^\top\boldsymbol{\delta}}$
    \EndFor
    \State $\Mmat^{(t+1)} = \text{diag}(\boldsymbol{\delta})\Qmat\text{diag}(\boldsymbol{\sigma})$
\EndFor
% \State $\Lcal_{\text{ot}} = \langle \Tmat, \Cmat \rangle$
\State \textbf{Return} $\langle \Mmat, \Cmat \rangle$
\end{algorithmic}
\end{algorithm}
% \end{minipage}
% \end{wrapfigure}
.

\subsection{Neural Machine Translation Experiments}
\label{ap_subsec: NMT}
\paragraph{Dataset} Two standard datasets are tested for NMT tasks. The first one is a small-scale English-Vietnamese corpus from the IWSLT 2015 Evaluation Campaign \citep{cettolo2015iwslt}, which is a parallel corpus of TED-talks and contains 133K sentence pairs. We follow the pre-processing procedure in \citep{luong2015stanford} by replacing words with frequencies less than 5 with $ \left \langle \text{unk} \right \rangle$. As a result, our vocabulary reduces to 17K for English and 7.7K for Vietnamese. We use TED tst2012 as development set and TED tst2013 as the test set.  For a large-scale dataset, we select an English-German corpus from the WMT16 Evaluate Campaign\footnote{\url{http://statmt.org/wmt16}}, which contains $4.5$M sentence pairs. Newstest 2013 is used as the development set and Newstest 2015 is used as the test set. We conduct the sub-word tokenization on the corpus using the Byte Pair Encoding (BPE) method \citep{sennrich2015neural}. Following \citet{opennmt}, we set the vocabulary size of both English and German to $32$K. 

\paragraph{Setup} We use Google's Neural Machine Translation (GNMT) system \citep{wu2016google} as our baseline MLE model, which follows the standard architecture and hyper-parameters\footnote{\url{https://github.com/tensorflow/nmt}} for fair comparison. All other models are built on top of with same network structure. We evaluate the model performance using BLEU scores \citep{papineni2002bleu}. We set OT weighting parameter $\lambda=0.1$ and order-preserving penalty weighting parameter $\beta=0.1$. 

For English-Vietnamese translation tasks (\ie, EN-VI or VI-EN), we follow the setup in \citep{sutskever2014sequence, luong2014addressing, luong2015effective}. We use one bidirectional LSTM layer with 512 hidden units as encoder and two-layer LSTM with 512 hidden units at each layer as decoder. The embedding dimension is set as 512. We follow the attention method described in \citep{luong2015effective} and use dropout with probability 0.2 as suggested by \citep{zaremba2014recurrent}. All parameters are initialized uniformly between $[-0.1, 0.1]$. We train the model for 12 epochs with 12 epochs using Stochastic Gradient Decent (SGD). For the first 8 epochs, we set learning rate as 1.0. After that, we anneal the learning rate at half at every epoch.

For English-German translation tasks (\ie,  EN-GE or GE-EN), we adopt a stacked LSTM with a 2-layer bidirectional of 1024 units as encoder and 4-layer LSTM with units 1024 as decoder. The embedding dimension is set to 1024. We adopt the attention used in \citep{wu2016google}. We train the model for 10 epochs. For the first 5 epochs, we set the learning rate as 1 and then halving the learning rate every half epoch.
% \subsection{Neural Machine Translation Experiment Dataset}

% \subsection{Neural Machine Translation Experiment Setup}
% \label{ap:setup}
% For English-Vietnamese translation tasks (\ie, EN-VI or VI-EN), we follow the setup in \citep{sutskever2014sequence, luong2014addressing, luong2015effective}. We use one bidirectional LSTM layer with 512 hidden units as encoder and two-layer LSTM with 512 hidden units at each layer as decoder. The embedding dimension is set as 512. We follow the attention method described in \citep{luong2015effective} and use dropout with probability 0.2 as suggested by \citep{zaremba2014recurrent}. All parameters are initialized uniformly between $[-0.1, 0.1]$. We train the model for 12 epochs with 12 epochs using Stochastic Gradient Decent (SGD). For the first 8 epochs, we set learning rate as 1.0. After that, we anneal the learning rate at half at every epoch.

% For English-German translation tasks (\ie,  EN-GE or GE-EN), we adopt a stacked LSTM with a 2-layer bidirectional of 1024 units as encoder and 4-layer LSTM with units 1024 as decoder. The embedding dimension is set to 1024. We adopt the attention used in \citep{wu2016google}. We train the model for 10 epochs. For the first 5 epochs, we set the learning rate as 1 and then halving the learning rate every half epoch. 
% % \section{Supplemental Material}

\subsection{Abstractive Text Summarization Experiments}
\label{ap_subsec:TS}
 We use a widely accepted English Gigawords corpus \citep{graff2003english} for the text summarization task. We follow the pre-process in \citep{rush2015neural}. The dataset is sampled and split into train/dev/test set with size 200K/8K/2K.

\subsection{Natural Language Generation Experiment}
\label{ap_subsec:NLG}
% \label{subsec:NLG_ex_setup}
In the NLG experiment, 200K sentences are collected as the training set and 10K sentences as the test set.
In the NLG experiment, we set OT weighting parameter $\lambda=1$ and the order-preserving penalty weighting parameter is $\beta=0.1$. Since input sequence $\yv$ is empty in a language model, to better guide the student forcing output, we adopt schedule sampling with ratio $0.3$ in our experiments. The samples generated by SFOT are presented in Table \ref{tab:SFOT_NLG_example}
\begin{table} [h] \scriptsize 
	\centering
	% \vskip 0.0in
	% \def\arraystretch{1.0}
	% \begin{scriptsize}
	\begin{tcolorbox} 
	\hspace{-5mm}
	\begin{tabular}{l}
% 		\toprule[1.2pt]
        % In addition , it was not clear that she would have been arrested by a police officer , the police said . \\
        So , this is a great way , but I'm not sure how to do that , he said . \\ \\
        When made an emergency landing , the driver who was also injured in the \\ blast was arrested on suspicion of causing death . \\ \\
        % \hline
        % We need to understand the vast majority of our staff , getting out of the villages , have been at home .\\
        % At the start of the day , I want to get the best possible deal for the next few years . \\
        
        % The announcement was released by the U . S . Supreme Court , which includes its right to take action against terrorism in the conflict .\\
        % The two of us were still in the car , but it ' s not a bad thing , he said .\\
        
        % The United Nations has launched a nuclear weapon to China during the conflict , which has been accused of taking an operation in the region .\\
        % Police have arrested the man on suspicion of murder , and prosecutors are trying to establish a crime . \\?
        The result is that the company's economic growth rate is rising by a \\ substantial margin in November, which is even higher than a year ago . \\ \\
        It's really a big deal for us and we're going to get ready for the \\second game .\\ \\
        
        We feel like it's very hard to say that it's really going to be the next \\ generation . \\ \\
        You don't want to be a kid , and there are a lot of things that you can do .\\ \\
        
        % And at the same time , the British government is facing a small number of tourists in Greece , France and Germany , and they will continue to stay in the EU .\\?
        
        % I want to be a coach who will play the best tennis and we ' re going to play and get better and it ' s just a big opportunity to get back and improving our game .\\
        
        The government's decision to extend its coal policy vote will be\\ announced in the first half of 2017 .\\ \\
        I'm not able to do it , but I think it's pretty important for him to be the\\ best player . \\ \\
        But I'm not sure what the supporters can do in this election , he said , \\referring to the Sanders campaign . \\
        % The Russian embassy has said the couple will be affected by a event in Ankara ' s eastern section .\\
        % In 1998 , the senator spoke with an attorney general ' s office on Monday evening , but he declined to comment .\\
      \end{tabular} 
      \vspace{-1mm}
      	\end{tcolorbox}
      
	\vspace{-3mm}
	\caption{Examples generated by SFOT in NLG experiments}
	\label{tab:SFOT_NLG_example}
	% \end{scriptsize}
	\vspace{-3mm}
\end{table}.
\label{sec:supplemental}
\end{document}